\documentclass[runningheads]{llncs}
\usepackage{eccv}
\usepackage{eccvabbrv}
\usepackage{graphicx}
\usepackage{amsmath,amssymb} 
\usepackage{color}
\usepackage{hyperref}
\usepackage{algorithmic}
\usepackage{amsmath}
\usepackage{amsfonts}
\usepackage{booktabs}
\usepackage{wrapfig}
\usepackage{graphicx}
\usepackage{multirow}
\usepackage{booktabs}
\usepackage{algorithm}   
\usepackage{soul}
\usepackage{arydshln}
\usepackage{orcidlink}
\usepackage{pdfpages}
\newcommand{\g}{\color{gray}} 
\usepackage{hyperref}

\begin{document}
\pagestyle{headings}
\mainmatter
\def\ECCV16SubNumber{8003}  

\title{Diffusion Image Generation with Explicit Modeling of Data Manifold Geometry} 

\titlerunning{Manifold-Aware Image Diffusion Model}

\authorrunning{D.~Xue et al.}

\author{Duoduo Xue\inst{1}\orcidlink{0000-0002-6608-5700}, 
Zhiyu Zhu\inst{1,2}\orcidlink{0000-0002-0726-4522}\thanks{Corresponding authors.}, 
Junhui Hou\inst{1}\orcidlink{0000-0003-3431-2021}\protect\footnotemark[1]}
\institute{Department of Computer Science, City University of Hong Kong, China
\and City University of Hong Kong (Dongguan), China \\
\email{duoduxue@cityu.edu.hk, zhiyu.zhu@cityu-dg.edu.cn, jh.hou@cityu.edu.hk}}

\maketitle 

\begin{abstract}
Image generative models aim to sample data points from the underlying data manifold, a task that requires learning and decoding a dense, low-dimensional, and compact parameterization space. To achieve this, we propose the Data Manifold-aware Image diffusioN moDel (MIND), a novel framework that explicitly models manifold geometry by integrating discrete patch tokenization into the score function of a continuous diffusion model. This approach successfully leverages both the structural quantification capabilities of discrete tokens and the parallel generation flexibility of continuous diffusion. Moreover, we enable end-to-end differentiable training via a novel soft top-$k$ aggregation mechanism and introduce dual-branch high-frequency feature embedding layers to alleviate the spectral bias of transformer backbones on low-dimensional inputs. Furthermore, for inference, we design a multi-stage transition sampling scheme that dynamically adjusts the sampling scheme based on timestep. Extensive experiments on ImageNet 256$\times$256 demonstrate the effectiveness of MIND. After 80-epoch training, our base model achieves an FID of 22.73 without guidance, nearly halving the 43.47 FID of the vanilla DiT-B/2 baseline. The proposed method reduces FID by 15.95 and 9.06 on average compared with the baselines DiT and SiT, respectively. For image generation on ImageNet 256$\times$256 with guidance, the proposed MIND-B with only 130M parameters achieves an FID of 2.06, surpassing  LlamaGen-3B with 3.1B parameters. Our MIND-XL with 715M parameters further reduces the FID to 1.95. Our MIND introduces a fresh perspective on diffusion-based image generation, paving the way for future research and innovation in this community. The code is available here: \url{https://github.com/xddgit/MIND}.
\end{abstract}

\section{Introduction}

Deep generative models, encompassing generative adversarial networks, variational autoencoders, and normalizing flows, have fundamentally revolutionized image synthesis~\cite{goodfellow2014generative,karras2019style,9089305}. In recent years, continuous diffusion models and score-based generative models~\cite{song2019generative,ho2020denoising,song2021denoising,nichol2021improved,NEURIPS2021_49ad23d1,NEURIPS2022_a98846e9,baoanalytic}, particularly latent diffusion models~\cite{rombach2022high,10203078} and Diffusion Transformers (DiT)~\cite{peebles2023scalable,ma2024sit}, have become the standard for high-fidelity image generation. Despite their impressive performance, these models typically operate within unbounded Euclidean spaces with infinite continuous states. While theoretical analyses heavily rely on the low-dimensional manifold hypothesis—which posits that high-dimensional natural images reside on a low-dimensional topological structure~\cite{azangulov2024convergence,cui2025a,loaiza-ganem2024deep,pmlr-v202-chen23o,pope2021the,wang2025diffusion,pmlr-v238-tang24a}, existing continuous diffusion algorithms rarely model this geometric prior explicitly, aside from a few specialized Riemannian diffusion formulations~\cite{NEURIPS2022_105112d5,kumar2026learning}. Consequently, image generation with standard continuous diffusion models is efficient but struggles to leverage the data manifold geometry optimally. 

Parallel to continuous diffusion models, discrete generation paradigms, e.g., next-token prediction~\cite{sun2024autoregressive,liu2025detailflow} or masked diffusion models~\cite{9878676,you2025effective}, compress images into discrete tokens via vector quantization~\cite{Esser_2021_CVPR,van2017neural,zhu2024scaling,shi2025scalable,xiongautoregressive}. Both of the methods (intentionally or unintentionally) set a decoding order onto the image generation process, e.g., the 1D generation order for the next-token prediction or random order for the discrete diffusion, which is constructed based on the assumption of causality among different patches. However, different from the language data, the success of vision diffusion models~\cite{9878676,ho2020denoising,peebles2023scalable,zheng2026diffusion} has demonstrated that high-fidelity of generation relies on progressively refining each token in parallel, which also applies to consistency-like models~\cite{pmlr-v202-song23a}.   
Furthermore, image generation models under the scheme of next-token prediction cannot utilize global bidirectional information during early sampling stages due to the inherent sequential generation scheme~\cite{xiongautoregressive}.

Explicitly constraining continuous diffusion processes to a low-dimensional geometric manifold is highly desirable but fundamentally challenging. Projecting continuous latent features directly onto a strict topological bottleneck (e.g., a low-dimensional hypersphere surface) leads to catastrophic representation collapse. Because continuous autoencoders rely heavily on vector magnitude to encode structural information, the strict $\ell_2$-normalization inherent to hyperspherical manifolds destroys this information, resulting in severe perceptual distortion. In contrast, discrete token indices encode information categorically. When mapped to a low-dimensional manifold, discrete tokens rely entirely on angular separation rather than magnitude, demonstrating an innate immunity to bottleneck degradation. See more analysis in Section \ref{sec: image tokenization}.

Motivated by this insight, we propose a novel and general generative framework named data Manifold-aware Image diffusioN moDel (MIND), which explicitly models the data manifold by leveraging the representational robustness of discrete tokenization with the sampling flexibility of continuous diffusion. Specifically, we first discretize images using a pre-trained tokenizer. The discrete tokens are then mapped onto a continuous low-dimensional hypersphere surface, upon which the forward and reverse diffusion processes are formulated. To enable end-to-end differentiable training between the unnormalized network logits output and the hyperspherical continuous latent, we introduce a soft top-$k$ aggregation and projection mechanism. Furthermore, to alleviate the spectral bias when processing low-dimensional inputs, we introduce a dual-branch feature projection module. This module injects Random Fourier Features to capture high-frequency details, stabilized by a zero-initialized deep residual sinusoidal pathway. During inference, we propose a multi-stage transition sampling scheme that dynamically adjusts the sampling method based on timestep. 

We evaluate the proposed method on the ImageNet 256$\times$256 benchmark. Under a restricted computational budget of 80 training epochs, our MIND-B (130M parameters) achieves an FID of 22.73, nearly halving the 43.47 FID of the vanilla DiT-B/2 baseline. In summary, our main contributions are three-fold:
\begin{itemize}
    \item We analyze the representation collapse phenomenon when projecting continuous latent onto a low-dimensional manifold, demonstrating the fundamental superiority of discrete tokens for low-dimensional manifold modeling.
    \item We propose a general data manifold-aware image diffusion model explicitly constrained to a hypersphere manifold. It features a differentiable soft top-$k$ bridge for training and multi-stage transition sampling for inference.
    \item Extensive experiments demonstrate that MIND significantly improves generation quality, achieving average FID reductions of 15.95 and 9.06 over the DiT and SiT baselines, respectively, establishing it as a highly efficient generative paradigm. For image generation on ImageNet 256$\times$256 with guidance, the proposed MIND-B with only 130M parameters achieves an FID of 2.06, surpassing  LlamaGen-3B with 3.1B parameters. Our MIND-XL with 715M parameters further reduces the FID to 1.95.
\end{itemize}

\section{Related Work}
\noindent\textbf{Diffusion Models.}
Diffusion models have beaten GAN and achieved great success in image generation recently~\cite{song2019generative,NEURIPS2021_49ad23d1}. Start from image generation in pixel space such as DDPM~\cite{ho2020denoising} and DDIM~\cite{song2021denoising}, latent diffusion models train diffusion models in latent space with much lower dimension by encoding images with an autoencoder to get continuous latent and enable high-resolution generation, the neural backbone is implemented with a time-dependent UNet in~\cite{rombach2022high} and a transformer in DiT~\cite{peebles2023scalable}. A series of works is built on the pioneering work of DiT. Scalable Interpolant Transformers (SiT)~\cite{ma2024sit} improves DiT with an interpolant framework. REPresentation Alignment (REPA)~\cite{yu2025representation} aligns the diffusion model representation with the pretrained visual representation. REPA-E~\cite{leng2025repa} proposes end-to-end training of the variational autoencoder (VAE) and diffusion model based on representation-alignment loss. Representation autoencoders~\cite{zheng2026diffusion} propose to replace the VAE in DiT with the well-developed and pretrained representation encoders to obtain a semantically rich latent space. Riemannian flow matching with Jacobi regularization~\cite{kumar2026learning} proposes to perform diffusion in the feature space of representation learning and constrain the generative process to the manifold geodesics. Deco~\cite{ma2025deco} proposes to decouple the generation as generating high-frequency details and low-frequency semantics in pixel and latent spaces, respectively. To improve the efficiency of diffusion models, several works explore one-step or few-step generation under the frameworks of consistency model~\cite{pmlr-v202-song23a}, flow-matching~\cite{geng2025mean,geng2025improved,huang2026rmflow}, and distillation~\cite{salimans2022progressive}.

Conditional diffusion models on class labels are crucial to further improve the generation quality. Class information is incorporated into adaptive group normalization layers together with time step in~\cite{NEURIPS2021_49ad23d1}. Dhariwal~\emph{et al.}~\cite{NEURIPS2021_49ad23d1} proposed to train a classifier on noisy images and then use the gradient of the trained classifier to guide the generation model. The widely used classifier-free guidance~\cite{ho2021classifier} is proposed to avoid an extra classifier, which trains a single network to parameterize both the unconditional and conditional models and samples with the linear combination of the conditional and unconditional score estimations. Rombach~\emph{et al.}~\cite{rombach2022high} proposed flexible conditional image generation by mapping the embedded class label to the intermediate layers of UNet with a cross-attention layer. Karras~\emph{et al.} proposed autoguidance that guides the model with an inferior version of itself~\cite{karras2024guiding} rather than an unconditional model as in~\cite{ho2021classifier}, which performs better but needs to train an additional guiding model. However, the diffusion models operator in continuous space with infinite states and cannot incorporate the prior information of the data manifold.

\noindent\textbf{Image generation with Language Models.}
Parallel to diffusion models, popular language models, such as autoregressive language models, have been extended to image generation since the great success of large language models in natural language generation. Images are firstly tokenized to discrete tokens $\boldsymbol{k}$ based on a well-developed visual tokenizer~\cite{yu2024language}. The autoregressive language models~\cite{sun2024autoregressive} predict the next token $\boldsymbol{k}_i$ using previous tokens $\boldsymbol{k}_1,\boldsymbol{k}_2,\cdots, \boldsymbol{k}_{i-1}$ and conditional information $\boldsymbol{c}$. The training stage learns the categorical distribution $p_{\boldsymbol{\theta}}(\boldsymbol{k}_i|\boldsymbol{k}_{i-1},c)$ by minimizing the negative log-likelihood $\mathcal{L}(\boldsymbol{\theta})=-\sum_{i=1}^N \log p_{\boldsymbol{\theta}}(\boldsymbol{k}_i \vert \allowbreak \boldsymbol{k}_1, \allowbreak \boldsymbol{k}_2, \allowbreak \cdots, \allowbreak \boldsymbol{k}_{i-1}; \allowbreak \boldsymbol{c})$, where $\boldsymbol{\theta}$ is the model parameters. The inference stage generates discrete tokens sequentially based on next-token prediction. The masked language models~\cite{9878676} learns the distribution $p_{\boldsymbol{\theta}}(\boldsymbol{k}_i|\boldsymbol{k}_j,c)$ in the training stage, where mask $\boldsymbol{m}\in\{0,1\}, m_j=1,\forall j, m_i=0, \forall i$. Starts with a fully masked sequence, the inference stage alternatively samples the whole sequence with the given non-masked tokens and remasks the tokens with the lowest probability in decreasing mask ratio. eMIGM (effective and efficient Masked Image Generation Models)~\cite{you2025effective} unifies the masked diffusion models~\cite{pmlr-v235-lou24a,shi2024simplified} and masked image generation~\cite{9878676}, which systematically explores the design space of masked image generation models. More recently, the diffusion language models have combined the advantages of diffusion models and language generation, which can be generalized to image generation directly~\cite{NEURIPS2021_958c5305,sahoo2024simple} by tokenizing images into discrete tokens. However, these language models restrict the whole generation process to a finite discrete space and limit the generation ability.

Several works try to combine the advantages of diffusion models and language models. The Riemannian diffusion language model (RDLM)~\cite{jo2025RDLM} proposes a continuous diffusion model for language modeling and performs the diffusion process on the hypersphere surface, which is extended by RJF~\cite{kumar2026learning} to image generation in the feature space of representation learning. Our MIND is fundamentally different from RDLM~\cite{jo2025RDLM} and RJF~\cite{kumar2026learning}. Specifically, in terms of \textit{motivation}, RJF aids convergence in high-dimensional representation spaces, whereas we embed a low-dimensional manifold prior. Regarding the \textit{input space}, RJF is purely continuous, while we process both discrete tokens and continuous latents. Finally, concerning the \textit{trajectory}, RDLM and RJF require complex, approximated formulations to strictly constrain trajectories to the hypersphere, whereas we allow off-sphere trajectories, only requiring data points on the surface.

\noindent \textbf{Image Tokenization} is the first step for image generation with autoregressive and masked language models. Images can be represented as discrete variables at the pixel level, but this is redundant and has a quadratically increasing computational cost. Image tokenizers compress images into discrete tokens. The general pipeline maps the image to a latent feature map and then quantizes the continuous features as discrete codes. A representative work is VQ-GAN~\cite{Esser_2021_CVPR} that constructs a learned codebook and finds the nearest codebook entry for each spatial feature vector. VQ-GAN is built on the grounding study named VQ-VAE~\cite{van2017neural} and introduces discriminator loss into the training objective. A series of works have been proposed to improve the initialization rate and size of the codebook. VQGAN-LC~\cite{zhu2024scaling} initializes the codebook entry with the feature of the pretrained vision encoder, then keeps the static codebook and optimizes a projector. The Index Backpropagation Quantization (IBQ) updates all codebook embeddings leveraging a soft one-hot categorical distribution~\cite{shi2025scalable}. The scale-based tokenizer named VAR proposes a multi-scale VQ autoencoder that quantizes residuals of feature maps recursively~\cite{tian2024visual}. Recently, visual language models have also been incorporated into image tokenizers for extracting semantically rich features~\cite{zheng2025vision}. A complementary survey of image tokenizers can be referred to~\cite{xiongautoregressive}.

\section{Proposed Method}

\begin{figure}[t]
    \centering
    \includegraphics[width=\linewidth]{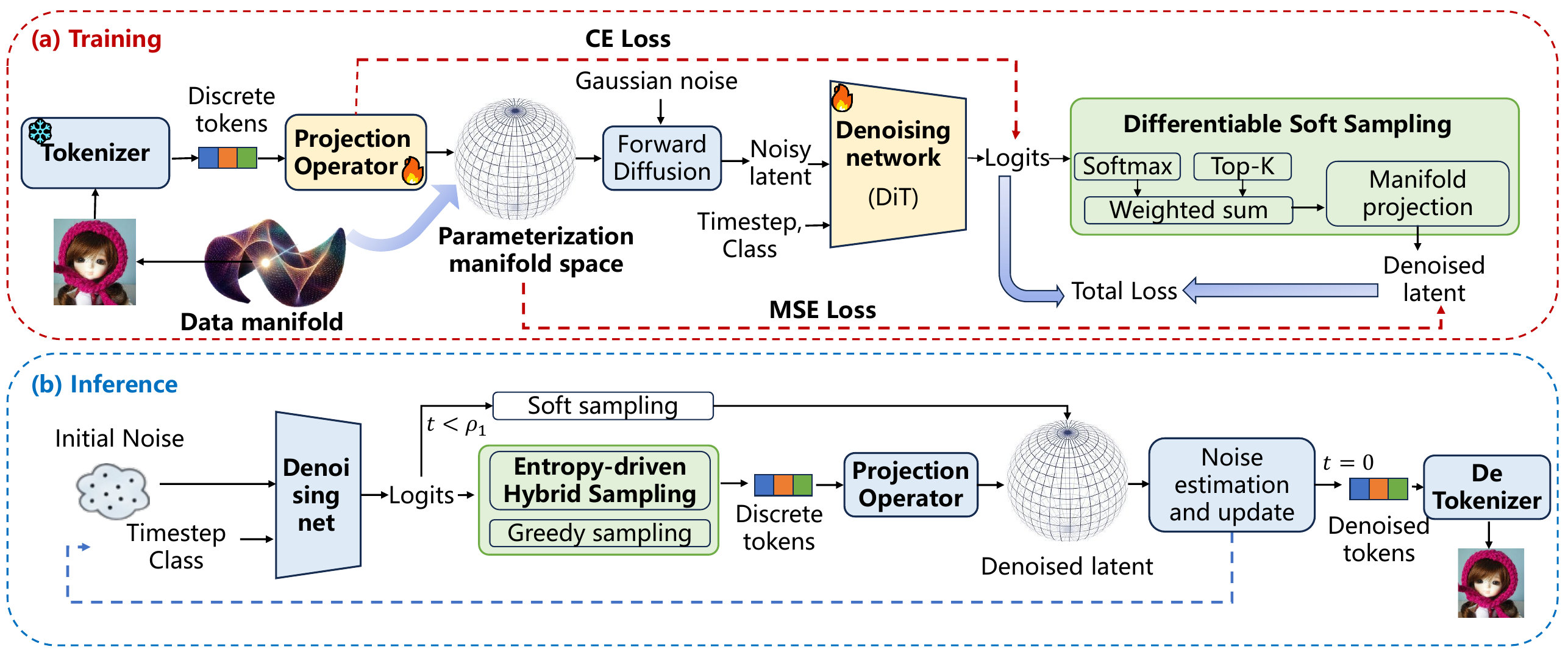} 
    \caption{Illustration of the proposed \textbf{MIND}, an image generation framework with explicit modeling of data manifold geometry.}
    \label{fig:placeholder}
\end{figure}

The low-dimensional manifold hypothesis is widely adopted in the theoretical analysis of image generation with diffusion models~\cite{azangulov2024convergence,loaiza-ganem2024deep,pmlr-v202-chen23o,pope2021the,wang2025diffusion,pmlr-v238-tang24a}, but has not been explicitly utilized in modern generative models. In this paper, we leverage the image tokenization method as the quantification measurement of image patch manifold structure and anchor those discrete points onto a simple and tackleable geometry, as illustrated in~\cref{fig:placeholder}. 

\subsection{Effectively Parameterizing the Image Manifold Geometry}
\label{sec: image tokenization}

Mapping images or their continuous latent representations onto a compact topological manifold presents a fundamental challenge, as capturing diverse visual information requires highly complex features. Furthermore, facilitating an effective generation process necessitates that real data samples densely populate the parameterization space. This strict high-density requirement severely exacerbates the difficulty of the learning objective.

To mitigate these challenges, we employ discrete tokenization to quantize the manifold structure, an approach that significantly reduces the reparameterization burden associated with continuous latent spaces. To validate this advantage, we empirically contrast two distinct methodologies: a \textit{continuous projection} that maps patchified VAE latents directly onto a $6$-dimensional hypersphere (where the exceptionally low dimensionality enforces a high-density space), and a \textit{discrete approach} that projects VQ token indices into the identical manifold.
\begin{wrapfigure}{r}{0.45\textwidth}
    \centering
    \includegraphics[width=\linewidth]{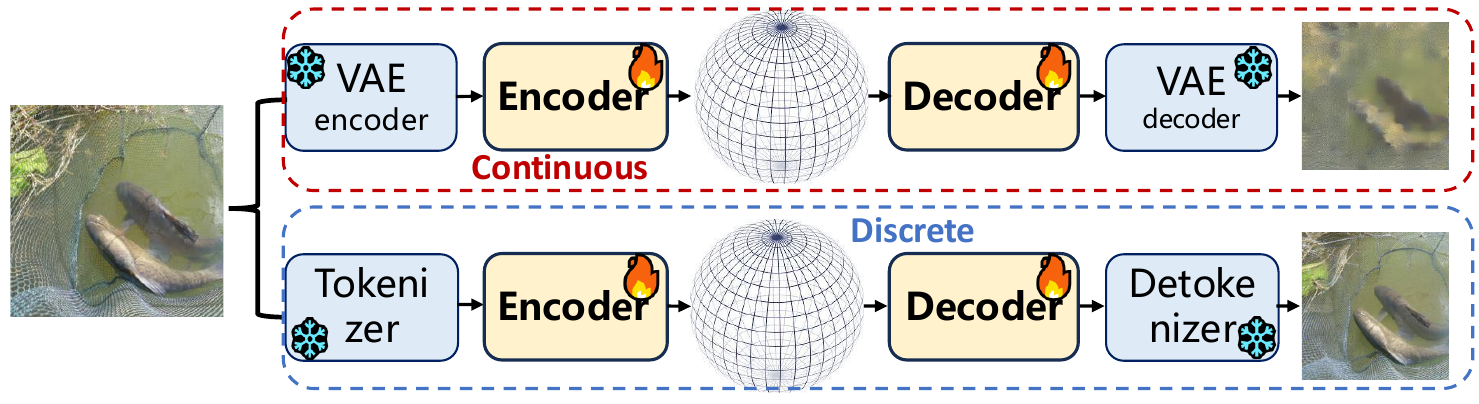}
    \includegraphics[width=\linewidth]{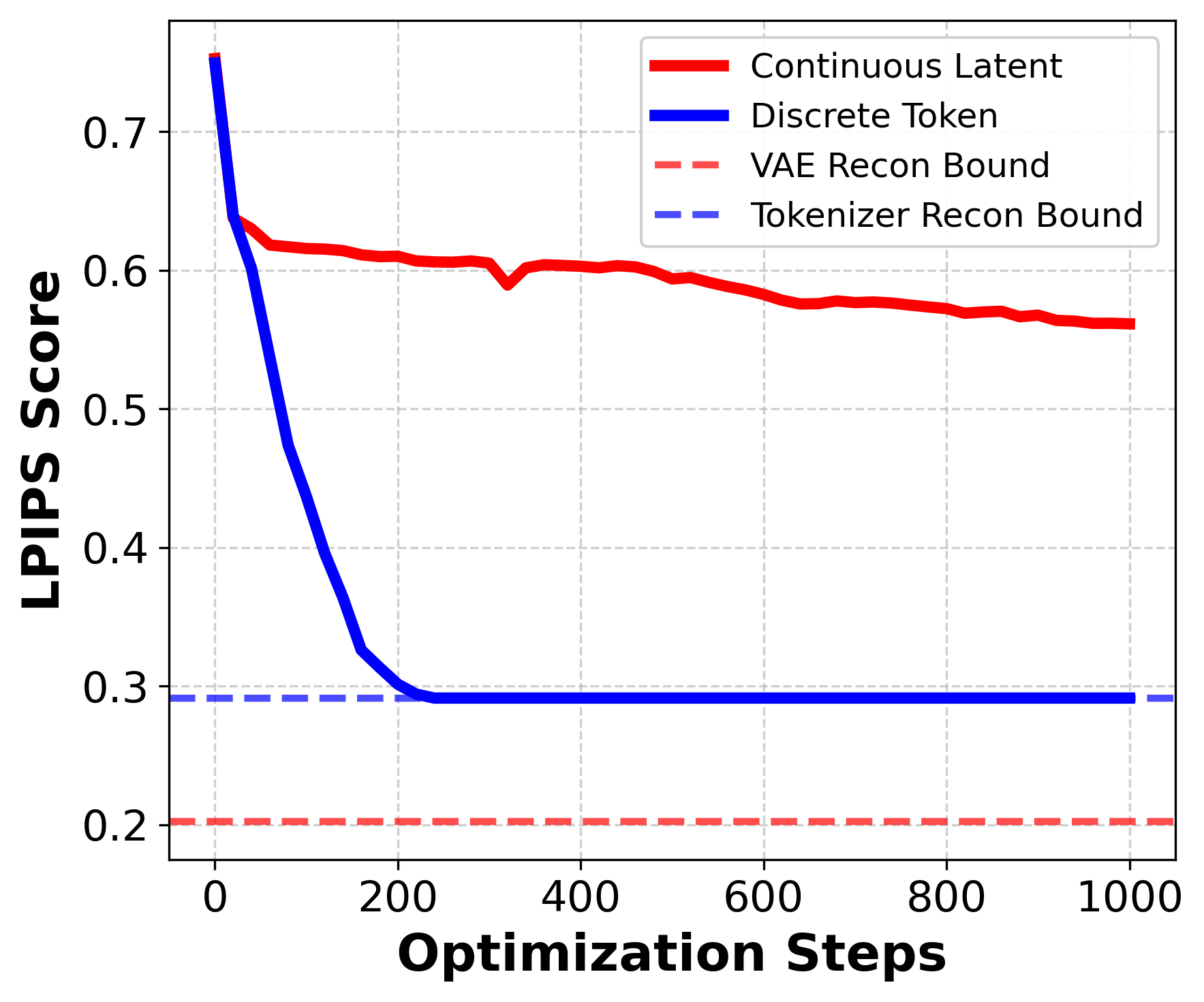}
    \caption{Continuous (top) and discrete projection (bottom). The plot shows the LPIPS distance between the reconstructed and original images during optimization. Solid lines represent the models with manifold constraints, whereas dashed lines indicate the unconstrained reconstruction.
    }
    \label{fig:fig1}
\end{wrapfigure}
For a rigorous and fair comparison, both methods are restricted to linear/MLP layers with the same parameter level and trained on the same randomly selected image, with carefully controlled optimization and architectural conditions (detailed in the \textit{Supplementary material}). By decoding the reconstructed bottleneck features back into the RGB domain, we measure the Learned Perceptual Image Patch Similarity (LPIPS)~\cite{zhang2018unreasonable}, as illustrated in~\cref{fig:fig1}. Our results indicate that the continuous latent formulation suffers from representation collapse due to its heavy reliance on feature magnitude to encode structural information. In contrast, the discrete token path encodes information categorically and relies entirely on angular separation. This grants it an innate robustness against the magnitude-destructive bottleneck, consistently yielding significantly lower perceptual distortion and demonstrating the inherent superiority of discrete tokens for manifold parameterization. Experiments with higher dimensions are shown in Fig.~S-5 of the \textit{Supplementary material}.

\subsection{Diffusion with Explicit Modeling of Data Manifold}

\noindent \textbf{Formulation of Diffusion Process.}
Given an image $\boldsymbol{I}\in\mathbb{R}^{C\times H\times W}$ sampled from data distribution $p_{data}$, it is tokenized as discrete token $\boldsymbol{k}\in\mathbb{R}^N$ using a tokenizer with vocabulary size $V$. The token $\boldsymbol{k}\in\mathbb{R}^N$ is then mapped to the continuous latent feature by a projection operator $\mathcal{P}_{\boldsymbol{\theta}}(\cdot)$ parameterized by $\boldsymbol{\theta}$ to get a continuous representation of dimension $L$ $\boldsymbol{x}_0\in\mathbb{R}^{N\times L}$ on the hypersphere surface satisfying $\sum_{l=0}^{L-1}\boldsymbol{x}^2_0(n,l)=R^2$ for all $n=0,\cdots,N-1$. 

The forward and reverse diffusion processes are performed based on the hypersphere surface with radius $R$. The forward diffusion process starts from the continuous representation $\boldsymbol{x}_0\in\mathbb{R}^{N\times L}$ and perturbs it to get noisy latent, 
\begin{equation}
\label{eq:forward_diffusion}
    \boldsymbol{x}_t = c_1 \sqrt{1-t}\cdot \boldsymbol{w}+ c_2 \sqrt{t}\cdot \boldsymbol{x}_0,
\end{equation}
where $\boldsymbol{w}\sim\mathcal{N}(\mathbf{0},\mathbf{I})$,$t\in[0,1]$, the end of forward diffusion process is the Gaussian noise $\boldsymbol{x}_1 = c_1 \boldsymbol{w}$. The reverse diffusion process starts from the sampled Gaussian noise $\boldsymbol{x}_1\sim\mathcal{N}(\mathbf{0},c_1^2\mathbf{I})$, the network $\boldsymbol{s}_{\boldsymbol{\phi}}(\cdot,t)$ parametrized by $\boldsymbol{\phi}$ gets noisy latent and outputs denoised logits $\tilde{\boldsymbol{x}}_{0t}=\boldsymbol{s}_{\boldsymbol{\phi}}(\boldsymbol{x}_t,t)\in\mathbb{R}^{N\times V}$. The denoised latent $\dot{\boldsymbol{x}}_{0t}$ is predicted by sampling discrete tokens from $\tilde{\boldsymbol{x}}_{0t}$ and then mapping the discrete tokens to continuous latent. 

\noindent\textbf{Training Pipeline.}
During the training stage, to maintain the differentiability of the discrete token sampling process while ensuring the output strictly resides on the target hypersphere manifold, we introduce a soft sampling mechanism. Let $\tilde{\boldsymbol{x}}_{0t}(n) \in \mathbb{R}^V$ denote the predicted logits of the $n$-th token over the vocabulary of size $V$. We first extract the set of indices corresponding to the top-$k$ logit values, denoted as $\mathcal{I}_k$. The normalized soft weights $\alpha_i$ are computed via the softmax function over these selected elements:
\begin{equation}
\label{eq:weight}
    \alpha_i = \exp(x_i)/\sum_{j \in \mathcal{I}_k} \exp(x_j), \forall i \in \mathcal{I}_k,
\end{equation}
where $x_i$ is the $i$-th element of $\tilde{\boldsymbol{x}}_{0t}(n)$. Subsequently, we aggregate the corresponding continuous latent using the weights $\alpha_i$. Since the convex combination of points on a hypersphere falls into the interior of the sphere, we explicitly project the aggregated feature onto the hypersphere surface via $\ell_2$-normalization:
\begin{equation}
\label{eq:soft sample}
    \dot{\boldsymbol{x}}_{0t}(n) = \frac{\sum_{i \in \mathcal{I}_k} \alpha_i \mathcal{P}_{\boldsymbol{\theta}}(i)}{\left\| \sum_{i \in \mathcal{I}_k} \alpha_i \mathcal{P}_{\boldsymbol{\theta}}(i) \right\|_2},
\end{equation}
where $\dot{\boldsymbol{x}}_{0t} \in \mathbb{R}^{N\times L}$ is the denoised continuous latent at timestep $t$, seamlessly bridging the discrete token space and the continuous diffusion manifold.

Based on the diffusion process formulated with the explicitly modeled hypersphere manifold prior, the neural network $\boldsymbol{s}_{\boldsymbol{\phi}}$ is optimized by minimizing the cross-entropy (CE) loss between denoised logits $\tilde{\boldsymbol{x}}_{0t}$ and the discrete tokens $\boldsymbol{k}$ and the mean squared error (MSE) loss between $\boldsymbol{x}_0$ and the denoised latent $\dot{\boldsymbol{x}}_{0t}$:
\begin{equation}
\label{eq:loss}
    \mathcal{L}(\boldsymbol{s}_{\boldsymbol{\boldsymbol{\phi}}},\mathcal{P}_{\boldsymbol{\theta}})=\mathbb{E}_{\boldsymbol{k}\sim p_{data}, t\in[0,1]} \{\mathrm{CE}(\tilde{\boldsymbol{x}}_{0t},\boldsymbol{k}) + \lambda \cdot \mathrm{MSE}(\dot{\boldsymbol{x}}_{0t},\boldsymbol{x}_0)\},
\end{equation}
where $\tilde{\boldsymbol{x}}_{0t}=\boldsymbol{s}_{\boldsymbol{\phi}}(\boldsymbol{x}_t,t)$, $\boldsymbol{x}_t = c_1 \sqrt{1-t}\cdot \boldsymbol{w}+ c_2 \sqrt{t}\cdot \boldsymbol{x}_0$, $\boldsymbol{x}_0=\mathcal{P}_{\boldsymbol{\theta}}(\boldsymbol{k})$, $\lambda$ controls the strength of MSE loss.~\Cref{alg:training} demonstrates the training pipeline of the proposed method.

\begin{algorithm}[t]
\caption{Training Process of MIND}
\label{alg:training}
\textbf{Input:} Dataset $p_{data}$, pre-trained tokenizer, initialized networks $\boldsymbol{s}_{\boldsymbol{\phi}}$ and $\mathcal{P}_{\boldsymbol{\theta}}$, noise schedules $c_1, c_2$, timestep range $[t_{min},t_{max}]$.
\begin{algorithmic}[1]
\REPEAT
    \STATE Sample image $\boldsymbol{I} \sim p_{data}$;
    \STATE Extract discrete tokens: $\boldsymbol{k} = \mathrm{Tokenizer}(\boldsymbol{I})$;
    \STATE Project to hypersphere manifold: $\boldsymbol{x}_0 = \mathcal{P}_{\boldsymbol{\theta}}(\boldsymbol{k})$;
    \STATE Sample timestep $t \sim \mathcal{U}[t_{min},t_{max}]$ and Gaussian noise $\boldsymbol{w} \sim \mathcal{N}(\mathbf{0}, \mathbf{I})$;
    \STATE Compute noisy latent: $\boldsymbol{x}_t = c_1 \sqrt{1-t}\cdot \boldsymbol{w} + c_2 \sqrt{t}\cdot \boldsymbol{x}_0$;
    \STATE Predict unnormalized logits: $\tilde{\boldsymbol{x}}_{0t} = \boldsymbol{s}_{\boldsymbol{\phi}}(\boldsymbol{x}_t, t)$;
    \STATE Obtain denoised latent $\dot{\boldsymbol{x}}_{0t}$ via soft sampling over $\tilde{\boldsymbol{x}}_{0t}$;
    \STATE Compute loss: $\mathcal{L}(\boldsymbol{s}_{\boldsymbol{\phi}},\mathcal{P}_{\boldsymbol{\theta}}) = \mathrm{CE}(\tilde{\boldsymbol{x}}_{0t}, \boldsymbol{k}) + \mathrm{MSE}(\dot{\boldsymbol{x}}_{0t}, \boldsymbol{x}_0)$;
    \STATE Take gradient descent step on $\nabla_{\boldsymbol{\theta},\boldsymbol{\phi}}\mathcal{L}$ and update $\boldsymbol{\theta},\boldsymbol{\phi}$;
\UNTIL{network converges}
\end{algorithmic}
\end{algorithm}

\begin{algorithm}[t]
\caption{Inference Process of MIND}
\label{alg:inference}
\textbf{Input:} Trained networks $\boldsymbol{s}_{\boldsymbol{\phi}}$ and $\mathcal{P}_{\boldsymbol{\theta}}$, noise schedules $c_1, c_2$, variance control coefficient $\eta\!\in\![0,\!1]$, sampling timestep schedule, thresholds $\rho_1,\rho_2$.
\begin{algorithmic}[1]
\STATE Sample initial noise: $\boldsymbol{x}_0 \sim \mathcal{N}(\mathbf{0}, c_1^2\mathbf{I})$;
\FOR{$t$=0 \textbf{to} 1}
    \STATE Predict unnormalized logits: $\tilde{\boldsymbol{x}}_{0t} = \boldsymbol{s}_{\boldsymbol{\phi}}(\boldsymbol{x}_t, t)$;
    \IF{$t<\rho_1$} \STATE $\dot{\boldsymbol{x}}_{0t}$ $\gets$ Soft sampling via Eq.~\eqref{eq:soft sample};
    \ELSE
    \STATE Sample discrete tokens with~\Cref{alg:sampling_operator}: $\hat{\boldsymbol{k}}_t = \mathrm{S}(\tilde{\boldsymbol{x}}_{0t})$;
    \STATE Project back to hypersphere manifold: $\dot{\boldsymbol{x}}_{0t} = \mathcal{P}_{\boldsymbol{\theta}}(\hat{\boldsymbol{k}}_t)$;
    \ENDIF
    \STATE Estimate noise component: $\boldsymbol{w}_{\boldsymbol{\phi}}(\boldsymbol{x}_{t},t) = (\boldsymbol{x}_t - c_2\sqrt{t} \cdot \dot{\boldsymbol{x}}_{0t})/(c_1\sqrt{1-t})$;
    \STATE Sample stochastic noise $\boldsymbol{w} \sim \mathcal{N}(\mathbf{0}, \mathbf{I})$;
    \STATE Compute latent for the next timestep: 
    \STATE \quad $\boldsymbol{x}_{t+\Delta t} = c_2\sqrt{t+\Delta t}\cdot\dot{\boldsymbol{x}}_{0t} + c_1\sqrt{1-(t+\Delta t)}\cdot \big(\eta \boldsymbol{w}_{\boldsymbol{\phi}}(\boldsymbol{x}_{t},t) + (1-\eta) \boldsymbol{w}\big)$;
\ENDFOR
\STATE \textbf{Return} DeTokenizer($\hat{\boldsymbol{k}}_1$).
\end{algorithmic}
\end{algorithm}

\noindent\textbf{Inference Pipeline.}
During the inference stage, the denoised logits are sampled to obtain the denoised latent  $\dot{\boldsymbol{x}}_{0t}=\mathcal{P}_{\boldsymbol{\theta}}(\mathrm{S}(\tilde{\boldsymbol{x}}_{0t}))$. To ensure that the generated token sequences remain in the high-probability manifold of the codebook space while preserving diversity in the early stage of the generation, we propose a parameterized multi-stage transition sampling strategy with thresholds $\rho_1$ and $\rho_2$. In the initial phase ($t<\rho_1$), the operator $\mathcal{P}(\mathrm{S}(\cdot))$ inherits the soft sampling mechanism from the training stage in Eq.~\eqref{eq:soft sample}. During the middle phase ($\rho_1<t<\rho_2$), the operator $\mathrm{S}(\cdot)$ is implemented by an entropy-aware discrete hybrid-filtering scheme. For the given distribution $\boldsymbol{p}=\mathrm{Softmax}(\tilde{\boldsymbol{x}}_{0t})$, we compute the Shannon entropy $H$. The sampling temperature is scaled adaptively: $\tau_{adj} = \tau \cdot (2.5e^{-H/3} + 0.6)$ following~\cite{matowards}, forcing the model to perform confident sampling when the predictive entropy is high. We then apply a hybrid filter combining Top-$k$ and Nucleus (Top-$p$) constraints, drawing the discrete token from the renormalized distribution. In the terminal phase ($t>\rho_2$), the operator $\mathrm{S}(\cdot)$ collapses to greedy sampling to eliminate stochasticity and ensure maximal local consistency. The detailed sampling scheme is summarized in~\Cref{alg:sampling_operator}. The denoised latent $\dot{\boldsymbol{x}}_{0t}\in\mathbb{R}^{N\times L}$ is perturbed to get the input of the next timestep, 
\begin{equation}
    \boldsymbol{x}_{t+\Delta t}=c_2\sqrt{t+\Delta t}\cdot\dot{\boldsymbol{x}}_{0t}+c_1\sqrt{1-(t+\Delta t)}\cdot (\eta \boldsymbol{w}_{\boldsymbol{\phi}}(\boldsymbol{x}_{t},t) + (1-\eta) \boldsymbol{w}),
\end{equation}
where $\boldsymbol{w}_{\boldsymbol{\phi}}(\boldsymbol{x}_{t},t)=(\boldsymbol{x}_t - c_2\sqrt{t} \cdot \dot{\boldsymbol{x}}_{0t})/(c_1\sqrt{1-t})$, $\eta\in[0,1]$.~\Cref{alg:inference} demonstrates the inference pipeline of the proposed method.

\begin{algorithm}[t]
\caption{Multi-Stage Transition Sampling Operator $\mathrm{S}(\cdot)$}
\label{alg:sampling_operator}
\begin{algorithmic}[1]
\STATE \textbf{Input:} Logits $\tilde{\boldsymbol{x}}_{0t}$, $t \in [0, 1]$, thresholds $\rho_1, \rho_2$, base temperature $\tau$, params $p, k$.
\STATE \textbf{Output:} Sampled token $\hat{\boldsymbol{k}}_t$.

\IF{$\rho_1\leq t<\rho_2$}
    \STATE \COMMENT{\textbf{Entropy-Driven Hybrid Sampling}}
    \STATE Predict entropy: $\boldsymbol{p} \gets \mathrm{Softmax}(\tilde{\boldsymbol{x}}_{0t})$, $H \gets - \sum \boldsymbol{p} \log \boldsymbol{p}$;
    \STATE Adaptive temperature: $\tau_{adj} \gets \tau \cdot (2.5 \cdot e^{-H/3} + 0.6)$;
    
    \STATE $\mathcal{V} \gets \mathrm{TopK}(\tilde{\boldsymbol{x}}_{0t}/\tau_{adj}, k) \cap \mathrm{Nucleus}(\boldsymbol{p}, p)$; 
    \STATE $\hat{\boldsymbol{k}}_t \sim \mathrm{Multinomial}\left(\mathrm{Softmax}(\tilde{\boldsymbol{x}}_{0t}/\tau_{adj}|{\mathcal{V}})\right)$;
\ELSIF{$t\geq \rho_2$}
    \STATE \textbf{Greedy sampling:} $\hat{\boldsymbol{k}}_t \gets \arg\max(\tilde{\boldsymbol{x}}_{0t})$;
\ENDIF
\RETURN $\hat{\boldsymbol{k}}_t$.
\end{algorithmic}
\end{algorithm}

\noindent\textbf{Network Architecture.}
The projector operator $\mathcal{P}_{\boldsymbol{\theta}}$ extracts the continuous latent $\boldsymbol{x}_0 \in \mathbb{R}^{N\times L}$ via an embedding layer, where each $L$-dimensional vector is factored into $d_{sub}$-dimensional hyperspherical subspaces through sub-vector normalization. The geometrically constrained $\boldsymbol{x}_0$ is then perturbed to yield the noisy latent $\boldsymbol{x}_t$. The denoising network $\boldsymbol{s}_{\boldsymbol{\phi}}$ is implemented based on the DiT. Before feeding into the DiT backbone, the continuous latent is processed through two branches named the base high-frequency mapper and the residual sinusoidal projection module. The high-frequency mapper maps the input $\boldsymbol{x}_t$ using a learnable projection matrix $\boldsymbol{B} \in \mathbb{R}^{L \times d_{map}}$, which is initialized from a Gaussian distribution $\mathcal{N}(0, \sigma^2)$ with a large variance to amplify high-frequency signals. The projected features are then transformed by sine and cosine functions,
\begin{equation}
    \boldsymbol{f}_{base} = \left[ \sin(2\pi \boldsymbol{x}_t \boldsymbol{B}), \cos(2\pi \boldsymbol{x}_t \boldsymbol{B}) \right].
\end{equation}
The concatenated features $\boldsymbol{f}_{base} \in \mathbb{R}^{2 d_{map}}$ are subsequently passed through a shallow Multi-Layer Perceptron (MLP) with Layer Normalization (LN) and GELU activations to yield the base hidden states $\boldsymbol{h}_{base} \in \mathbb{R}^{d_{hidden}}$, matching the hidden dimension of the DiT backbone. The residual sinusoidal projection module expands the continuous latent $\boldsymbol{x}_t$ using a deterministic sinusoidal positional encoding to get a high-dimensional vector $\boldsymbol{f}_{freq}$, which is passed through an MLP equipped with LN and SiLU activations. Crucially, to ensure stable gradient flow and prevent catastrophic divergence at initialization, the weight matrix of the final linear layer in this residual MLP is strictly initialized to zero. Finally, the outputs from both branches are aggregated via a residual addition $\boldsymbol{h}_{input} = \boldsymbol{h}_{base} + \boldsymbol{h}_{res}$ before interacting with the timestep and class conditions. The backbone network utilizes the transformer-based diffusion architecture in DiT, which scales across multiple model capacities. Conditioning is handled via an adaptive LayerNorm that injects combined timestep and class label representations into each Transformer block, while spatial dependencies are modeled using rotary positional embeddings. 
\section{Experiments}
\subsection{Experimental Settings} 
To evaluate our proposed method, we followed the setup of DiT~\cite{peebles2023scalable} and SiT~\cite{ma2024sit} to perform experiments on the standard ImageNet dataset~\cite{5206848}. Each image was processed to the resolution of $256\times 256$ and encoded into discrete token sequences of length 256 using the pre-trained tokenizers Index Backpropagation Quantization (IBQ)~\cite{shi2025scalable} or GigaTok~\cite{gigatok} with a vocabulary size of $V$. These discrete tokens are mapped to continuous latent features on the hypersphere surface. The model was trained within the continuous feature space using the forward noising formulation in Eq.~\eqref{eq:forward_diffusion} and the hybrid loss combining cross-entropy and MSE defined in Eq.~\eqref{eq:loss}. To enable classifier-free guidance during inference, we randomly replaced class labels with an unconditional token with a probability of $p_{drop} = 0.1$ during training. All models were optimized using the AdamW optimizer, utilizing chunked gradient checkpointing to optimize memory consumption. Detailed training hyperparameters, including specific learning rates and batch sizes for each experimental setting, are comprehensively summarized in Table~S-8 of the \textit{Supplementary material}. During the inference phase, we employed a custom SDE solver in~\Cref{alg:inference} with 250 denoising steps to iteratively map Gaussian noise back to the data manifold. At each timestep $t$, the backbone network $\boldsymbol{s}_{\boldsymbol{\phi}}$ processed the continuous latent state to predict the unnormalized logit distribution over the discrete token vocabulary. We applied classifier-free guidance with scale $cfg$ for condition alignment. 

\subsection{Performance Evaluation}
We report standard generative metrics: Fréchet Inception Distance (FID), Inception Score (IS), Precision, and Recall on 50,000 samples that are generated with class-balanced sampling following~\cite{zheng2026diffusion}.

\noindent\textbf{Image Generation without Guidance.}
We performed a comprehensive system-level evaluation comparing our proposed model against the vanilla DiT baseline. To ensure a fair comparison under a constrained computational budget, all models were evaluated after training for 80 epochs on the ImageNet 256$\times$256 dataset without classifier-free guidance. Table~S-8 of the \textit{Supplementary material} summarizes the hyperparameters during inference. As shown in~\Cref{tab:uncondtional}, the proposed MIND significantly outperforms the vanilla DiT, demonstrating vastly superior performance. Specifically, at the small (S) scale, the proposed MIND-S achieves a remarkable FID of 40.72, yielding a massive absolute improvement of 27.68 over DiT-S/2 (68.40). This trend is further amplified at the base (B) scale, where the proposed MIND-B reaches an FID of 22.73, nearly halving the FID of DiT-B/2 (43.47). Furthermore, compared to more advanced frameworks such as the flow-based SiT and the discrete diffusion-based model eMIGM~\cite{you2025effective}, the proposed MIND establishes state-of-the-art FID while maintaining highly competitive diversity. It should be noted that the proposed method using 130M parameters even outperforms DiT-L with 458M parameters. 
\begin{table}[t]
    \centering
    \caption{\textbf{Class-conditional image generation without guidance on Imagenet 256$\times$ 256.} Bold values indicate the best performance. ``N/A'' denotes that the metric was not reported in the original publication. All models are trained for 80 epochs.}
    \label{tab:uncondtional}
    \setlength{\tabcolsep}{4pt}
    \begin{tabular}{@{}lcccccc@{}}
        \toprule
        &\#Params & FID $\downarrow$ & IS $\uparrow$ & Precision $\uparrow$ & Recall $\uparrow$ \\
        \midrule
        DiT-S/2~\cite{peebles2023scalable} &33M &68.40 &N/A &N/A &N/A\\ 
        SiT-S/2~\cite{ma2024sit} &33M &57.64 &24.78 &0.41 &0.60\\
        MIND-S(Our)&$\sim$35M &\textbf{40.72} &\textbf{31.51} &\textbf{0.48} &\textbf{0.61} \\
        \midrule 
        eMIGM-S~\cite{you2025effective} &97M &44.47&19.82&0.49&0.57\\
        \midrule
        DiT-B/2~\cite{peebles2023scalable} &130M &43.47 &N/A &N/A &N/A\\
        SiT-B/2~\cite{ma2024sit} &130M & 33.02 &43.71 &0.53 &\textbf{0.63}\\ 
        MIND-B(Our) &$\sim$ 130M &\textbf{22.73} &\textbf{56.17}&\textbf{0.55}&0.58\\
        \midrule
        DiT-L/2~\cite{peebles2023scalable} &458M &23.33&N/A &N/A &N/A\\ 
        \bottomrule  
    \end{tabular}
\end{table}

\begin{table}[ht]
    \centering
    \caption{\textbf{Class-conditional image generation with guidance on ImageNet 256$\times$256.} Bold values indicate the best performance. All models were trained for 80 epochs. Our method achieves a superior FID of 7.97, outperforming the baseline DiT (10.95) and flow-based SiT (9.07).}
    \label{tab:condtional}
    \setlength{\tabcolsep}{4pt}
    \begin{tabular}{@{}lcccccc@{}}
        \toprule
        &\#Params &$cfg$& FID $\downarrow$ & IS $\uparrow$ & Precision $\uparrow$ & Recall $\uparrow$ \\
        \midrule
        DiT-S/2~\cite{peebles2023scalable}&33M &1.5 &44.42&34.57&0.47&0.56\\
        SiT-S/2~\cite{ma2024sit} &33M &1.5&35.62&43.00&0.52&\textbf{0.56}\\
        MIND-S(Our)&$\sim$35M &1.5&
        \textbf{22.33}&\textbf{60.32}&\textbf{0.63}&0.52\\
        \hdashline
        DiT-S/2~\cite{peebles2023scalable} &33M &2.0&29.29&55.37&0.57&0.51\\
        SiT-S/2~\cite{ma2024sit} &33M &2.0&22.96&67.81& 0.62&\textbf{0.51}\\
        MIND-S(Our)&$\sim$35M &2.0 &\textbf{14.93}&\textbf{92.65}&\textbf{0.74}&0.45\\
        \midrule 
        eMIGM-S~\cite{you2025effective} &97M &1.3&41.18&21.23& 0.50&0.56\\
        eMIGM-S~\cite{you2025effective} &97M &4.0&23.52&30.83 &0.61 &0.51\\
        \midrule
        DiT-B/2~\cite{peebles2023scalable}&130M &1.5&20.01&73.00&0.65&0.56\\
        SiT-B/2~\cite{ma2024sit} &130M &1.5&16.88&84.26&0.66&\textbf{0.56}\\
        MIND-B(Our) &$\sim$ 130M &1.5&\textbf{12.15}&\textbf{100.16}&\textbf{0.71}&0.51\\
        \hdashline
        DiT-B/2~\cite{peebles2023scalable}&130M&2.0 &10.95&119.12&0.76&0.47\\
        SiT-B/2~\cite{ma2024sit} &130M &2.0&9.07&137.07&0.77&\textbf{0.47}\\
        MIND-B(Our) &$\sim$ 130M &2.0&\textbf{7.97}&\textbf{154.20}&\textbf{0.81}&0.43\\
        \bottomrule
    \end{tabular}
\end{table}

\noindent\textbf{Image Generation with Guidance.} \Cref{tab:condtional} presents the quantitative results for class-conditional image generation with guidance on ImageNet $256\times256$. 
\begin{table*}[!tbp]
\centering
\caption{Quantitative comparison of image generation models on ImageNet $256 \times 256$. With only $\sim$130M parameters, our method achieves performance comparable to or exceeding discrete models with billions of parameters and Continuous models with hundreds of millions of parameters. ``-re'': rejection sampling, ``-G'': GigaTok, $^*$: taken from VAR~\cite{tian2024visual}.}
\label{tab:sota_comparison}
{
\begin{tabular}{lccccccc}
\toprule
\textbf{Method} & \textbf{Family} & \textbf{\#Params} & \textbf{Epochs} & \textbf{FID} $\downarrow$ & \textbf{IS} $\uparrow$ & \textbf{Prec.} $\uparrow$ & \textbf{Rec.} $\uparrow$ \\
\midrule
\multicolumn{8}{c}{\textit{Discrete Models}} \\
\midrule
VQGAN~\cite{Esser_2021_CVPR} & AR &1.4B &N/A &15.78 &74.3 &N/A &N/A\\
VQGAN-re~\cite{Esser_2021_CVPR} & AR &1.4B &N/A & 5.20 &280.3 &N/A &N/A\\
ViTVQ~\cite{yu2022vectorquantized} &AR &1.7B&360 &4.17& 175.1&N/A &N/A\\
ViTVQ-re~\cite{yu2022vectorquantized} & AR &1.7B&360 &3.04& 227.4&N/A &N/A\\
RQTran.~\cite{lee2022autoregressive}&AR &3.8B &N/A &7.55 &134.0&N/A &N/A\\
RQTran.-re~\cite{lee2022autoregressive}&AR &3.8B &N/A &3.80 &323.7&N/A &N/A\\
LlamaGen-B~\cite{sun2024autoregressive} &AR &111M &300 &5.46 &193.61 &0.83 &0.45 \\
LlamaGen-L~\cite{sun2024autoregressive} &AR &343M &300 &3.07 &256.06 &0.83 &0.52 \\
LlamaGen-XL~\cite{sun2024autoregressive} &AR &775M &300 &2.62 &244.08 &0.80 &0.57 \\
LlamaGen-XXL~\cite{sun2024autoregressive} &AR &1.4B &300 &2.34 &253.90 &0.80 &0.59 \\
LlamaGen-3B~\cite{sun2024autoregressive} & AR &3.1B &300 &2.18 &263.33 &0.81 &0.58\\
\midrule 
VAR-d16~\cite{tian2024visual} & VAR & 310M &200-350 & 3.30 &274.4 &0.84 &0.51\\
VAR-d20~\cite{tian2024visual} & VAR & 600M &200-350 & 2.57 &302.6 &0.83 &0.56 \\
VAR-d24~\cite{tian2024visual} & VAR & 1.0B &200-350 & 2.09 &312.9 &0.82 &0.59 \\
VAR-d30~\cite{tian2024visual} & VAR & 2.0B &200-350 & 1.92 &323.1 &0.82 &0.59\\
\midrule 
IBQ-B~\cite{shi2025scalable} &AR &342M &300 &2.88 &254.73 &0.84 &0.51 \\
IBQ-L~\cite{shi2025scalable} &AR&649M &350 &2.45 &267.48 &0.83 &0.52 \\
IBQ-XL~\cite{shi2025scalable} &AR &1.1B &400 &2.14 &278.99 &0.83 &0.56 \\
IBQ-XXL~\cite{shi2025scalable} &AR &2.1B &450 &2.05 &286.73 &0.83 &0.57 \\
\midrule 

MaskGIT~\cite{9878676} & Masked& 227M &300 & 6.18 & 182.1 & 0.80 & 0.51 \\
MaskGIT-re~\cite{9878676} & Masked & 227M &300 &4.02 &355.6 &N/A &N/A \\
\midrule
\textbf{MIND-B(Ours)} & \textbf{Diff.+Dis.} & \textbf{$\sim$130M} &896 & \textbf{2.21} &\textbf{260.54} &\textbf{0.80} &\textbf{0.58} \\
\textbf{MIND-B(Ours)} & \textbf{Diff.+Dis.} & \textbf{$\sim$130M} &1000 &\textbf{2.18} &\textbf{258.46} &\textbf{0.80} &\textbf{0.58}\\
\textbf{MIND-B-G(Ours)} & \textbf{Diff.+Dis.} & \textbf{$\sim$130M} &1600 &\textbf{2.06} &\textbf{268.03} &\textbf{0.78} &\textbf{0.62}\\
\textbf{MIND-XL(Ours)} & \textbf{Diff.+Dis.} & \textbf{$\sim$715M} &1000 & \textbf{1.97} &\textbf{297.93} &\textbf{0.78} &\textbf{0.61} \\
\textbf{MIND-XL-G(Ours)} & \textbf{Diff.+Dis.} & \textbf{$\sim$715M} & 1600& \textbf{1.95} &\textbf{293.78} &\textbf{0.75} &\textbf{0.67} \\
\midrule 
\multicolumn{8}{c}{\textit{Continuous Models}} \\
\midrule 
BigGAN~\cite{brock2018large}& GAN &112M &N/A& 6.95 &224.5 &0.89 &0.38 \\
GigaGAN~\cite{10205294} &GAN &569M &124 &3.45 &225.5& 0.84 &0.61\\
StyleGan-XL~\cite{10.1145/3528233.3530738} &GAN &166M&N/A & 2.30 &265.1 &0.78 &0.53 \\
\midrule
ADM-G~\cite{NEURIPS2021_49ad23d1}& Diff. &554M &400 &4.59 &186.70 &0.82 &0.52\\
ADM-G, ADM-U~\cite{NEURIPS2021_49ad23d1}& Diff. &554M &400 & 3.94 &215.84 &0.83 &0.53 \\
LDM-4-G~\cite{rombach2022high} & Diff. & 400M &167 &3.60 & 247.7 & 0.87 & 0.48 \\
DiT-L/2$^*$~\cite{peebles2023scalable} &Diff.&458M &1400 &5.02 &167.2 &0.75 &0.57 \\
DiT-XL/2~\cite{peebles2023scalable} & Diff. & 675M &1400 &2.27 & 278.2 & 0.83 & 0.57 \\
SimDiff~\cite{pmlr-v202-hoogeboom23a} &Diff. &2B &800 &2.77 &211.8 &N/A &N/A \\
SiT-XL/2~\cite{ma2024sit} & Diff. & 675M &1400 &2.06 & 270.3 & 0.82 & 0.59 \\
\midrule
\end{tabular}}
\end{table*}
\begin{table*}[!ht]
\ContinuedFloat
\centering
\caption{Quantitative comparison (continued). $^{**}$: Models used large-scale pre-trained representation network.}
\label{tab:sota_comparison-cont}
\begin{tabular}{lccccccc}
\toprule
\textbf{Method} & \textbf{Family} & \textbf{\#Params} & \textbf{Epochs} & \textbf{FID} $\downarrow$ & \textbf{IS} $\uparrow$ & \textbf{Prec.} $\uparrow$ & \textbf{Rec.} $\uparrow$ \\
\midrule 
\multicolumn{8}{c}{\textit{Continuous Models}} \\
\midrule
eMIGM-XS~\cite{you2025effective} &Masked &69M &800 &3.62 &224.91 &0.80 &0.51 \\
eMIGM-S~\cite{you2025effective} &Masked &97M &800 & 2.87 &254.48 &0.80 &0.54 \\
eMIGM-B~\cite{you2025effective} &Masked &208M &800 &2.32 &278.97 &0.81 &0.57 \\
eMIGM-L~\cite{you2025effective} &Masked &478M &800 &1.72 &304.16 &0.80 &0.60 \\
eMIGM-H~\cite{you2025effective} &Masked &942M &800 &1.57 &305.99 &0.80 &0.63 \\
\midrule
MAR-B~\cite{NEURIPS2024_66e22646}& MAR& 208M &800 &2.31 &281.7 &0.82 &0.57\\ 
MAR-L~\cite{NEURIPS2024_66e22646}& MAR& 479M &800 &1.78 &296.0 &0.81 &0.60 \\
MAR-H~\cite{NEURIPS2024_66e22646}& MAR& 943M &800 &1.55 &303.7 &0.81 &0.62 \\
\midrule
\g REPA$^{**}$~\cite{yu2025representation} &Diff. &675M &800 &1.29 &306.3 &0.79 &0.64 \\
\g REPA-E$^{**}$~\cite{leng2025repa} &Diff. &675M &800 &1.12 &302.9 &0.79 &0.66\\
\g DDT$^{**}$~\cite{wang2025ddt} &Diff. &675M &400 &1.26 &310.6 &0.79 &0.65\\
\g RAE$^{**}$~\cite{zheng2026diffusion} &Diff. &839M &800 &1.13 &262.6 &0.78 &0.67\\
\midrule
\g PixelFlow~\cite{chen2025pixelflow}& Flow &677M &320 &1.98 &282.1 &0.81 &0.60\\
\g MeanFlow-XL/2~\cite{geng2025mean} & Flow & 676M &1000 & 2.20 &N/A &N/A &N/A \\
\g IMF-XL/2~\cite{geng2025improved}&Flow &610M &800 &1.54 &N/A &N/A &N/A\\ 
\bottomrule
\end{tabular}
\end{table*}
In the absence of publicly available 80-epoch performance and checkpoints for DiT, SiT, and eMIGM, we report results by our training from scratch following the default hyperparameter configurations specified in the original works. We sampled with different classifier-free guidance scale $cfg$ and default hyperparameters in the original works. eMIGM was sampled with the recommended higher $cfg$ since it adopts the time interval guidance strategy by default. The proposed MIND demonstrates superior generative performance across different model scales and classifier-free guidance scales. Under the small-scale setting ($\sim$35M parameters), the proposed MIND-S significantly outperforms the baseline models. Specifically, at $cfg=1.5$, MIND-S improves the FID by 22.09 and 13.29 compared with DiT-S/2 and SiT-S/2, respectively. MIND-S surpasses eMIGM-S with only one-third of the parameters. Scaling up to the base configuration ($\sim$130M parameters), the proposed MIND-B consistently exhibits strong capabilities. At $cfg=1.5$, MIND-B reaches the best FID of 12.15 and the highest IS of 100.16. When the guidance scale is increased to 2.0, our model reduces FID by 2.98 and 1.1 compared with DiT-B/2 and SiT-B/2, respectively.

\begin{figure}[ht]
    \centering
    \includegraphics[width=0.8\linewidth]{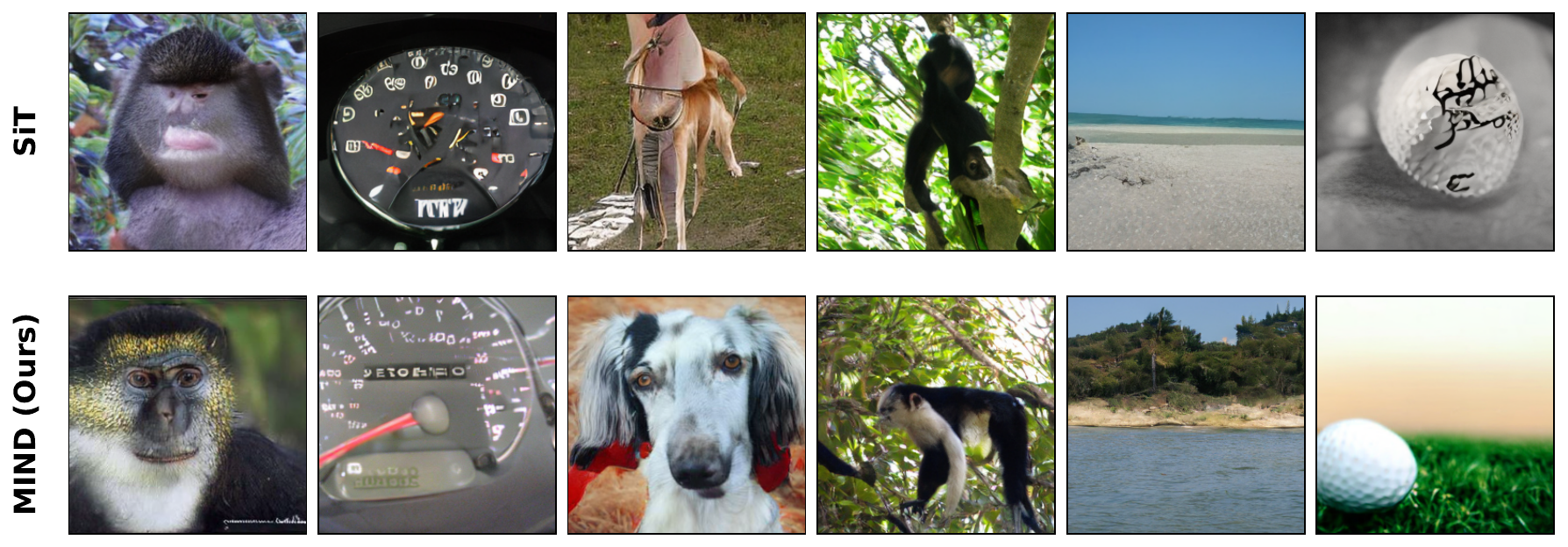} 
    \caption{Visual comparison between our MIND-B and SiT-B/2 with $cfg$=2.0.}
    \label{fig:visual compare}
\end{figure}
\cref{fig:visual compare} illustrates a qualitative comparison between SiT-B/2 and the proposed MIND-B, using the same class labels with $cfg=2.0$. Our method synthesizes significantly more realistic images with precise semantic alignment, effectively avoiding the structural distortions observed in SiT-B/2. These visual results align with our superior quantitative scores in FID, IS, and Precision, which demonstrate that prioritizing sample fidelity and precision over diversity (Recall) yields more visually compelling and structurally accurate generative outcomes. Fig.~S-4 of the \textit{Supplementary material} provides some visual results from our MIND-B with $cfg$=4.0. We believe that the performance of larger models will be further unlocked by increasing vocabulary size in future work.

\subsection{System-level Comparisons with Prior Work}
In this section, we present the ultimate performance evaluation of our model against existing state-of-the-art methods. For inference, the classifier-free guidance scale is applied in a limited interval without entropy-driven adaptive temperature as proposed in~\cite{nniemi2024applying}. The training and inference parameters remain consistent with Table~S-8 of the \textit{Supplementary material} except for those summarized in Table~S-9. MIND-B-G and MIND-XL(-G) are sampled with NPU-compatible code.

As shown in~\Cref{tab:sota_comparison}, our MIND with about 130M parameters achieves a highly competitive FID of 2.06 (IS: 268.03, Precision: 0.78), successfully outperforming mainstream baselines that are substantially larger across multiple generative paradigms. Specifically, MIND surpasses billion-parameter discrete models like LlamaGen-3B (3.1B, FID 2.18), RQTran.~\cite{lee2022autoregressive} (3.8B, FID 3.80), IBQ-XXL (2.1B, FID 2.05)~\cite{shi2025scalable}, VAR~\cite{tian2024visual} (1.0B, FID2.09). 

Furthermore, our method maintains a distinct advantage over continuous architectures. Specifically, compared with its baseline DiT-XL/2~\cite{peebles2023scalable}, MIND improves the generation ability from FID=2.27 to 1.95. The proposed MIND, using a stronger tokenizer named GigaTok~\cite{10205294}, outperforms the 2B-parameter SimDiff~\cite{pmlr-v202-hoogeboom23a} and SiT-XL/2~\cite{ma2024sit}. Compared with recent masked generative models that rely on continuous tokens, such as eMIGM-B~\cite{you2025effective} and MAR-B~\cite{NEURIPS2024_66e22646}, the proposed MIND achieves comparable performance. We anticipate that the proposed framework can achieve even greater performance by introducing a residual term to mitigate the information loss inherent in vector quantization. Ultimately, our approach challenges the conventional reliance on massive parameter scaling, demonstrating that a compact $\sim$130M model can effectively match or exceed the performance of state-of-the-art architectures ranging from 300M to 3B parameters. 

More visual results, ablation studies, and efficiency analysis are provided in the \textit{Supplementary material}.
\section{Conclusion and Discussion}
In this paper, we introduced a novel generative framework that explicitly models the data manifold by bridging discrete image tokenization with continuous hyperspherical diffusion. To enable this discrete-continuous hybrid system, we proposed a differentiable soft top-$k$ aggregation mechanism for stable training and an entropy-driven hybrid sampling operator for robust inference. Furthermore, our dual-branch high-frequency feature mapper effectively resolves the spectral bias when processing low-dimensional inputs. Extensive experiments on ImageNet demonstrate that our approach significantly outperforms the DiT baseline, flow-based continuous models, and discrete masked diffusion models. 

Note that while we instantiated our method on the basic DiT architecture to highlight its fundamental superiority, our significant gains over DiT show its immense potential for image generation. We will integrate advanced trajectory formulations, such as mean flow and consistency models, to enhance sampling efficiency and enable high-fidelity, few-step, or single-step generation. Additionally, incorporating self-supervised representation learning could further enrich feature expression and accelerate training convergence to get better performance.
\section*{Acknowledgements}
This work was supported in part by the National Natural Science Foundation of China under Grants 62422118, and in part by the Hong Kong Research Grants Council under Grants N\_CityU1114/25 and 11219324.
\bibliographystyle{splncs04}
\bibliography{egbib}
\includepdf[pages=-]{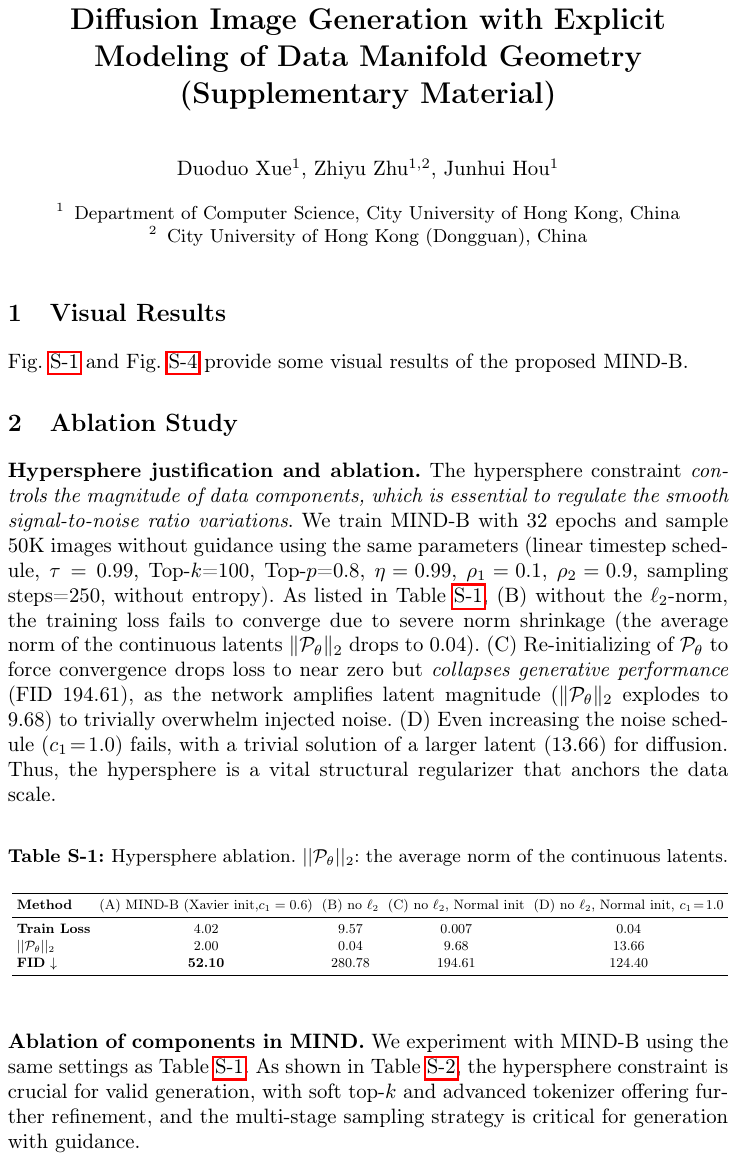}
\end{document}